\documentclass[10pt,twocolumn,letterpaper]{article}

\usepackage{cvpr}              

\usepackage[flushleft]{threeparttable}
\usepackage{graphicx}
\usepackage{amsmath}
\usepackage{amssymb}
\usepackage{booktabs}
\usepackage{multirow}

\usepackage[pagebackref,breaklinks,colorlinks]{hyperref}

\usepackage[capitalize]{cleveref}
\crefname{section}{Sec.}{Secs.}
\Crefname{section}{Section}{Sections}
\Crefname{table}{Table}{Tables}
\crefname{table}{Tab.}{Tabs.}

\def\cvprPaperID{*****} 
\def\confName{CVPR}
\def\confYear{2023}

\begin{document}

\title{FedBKD: Distilled Federated Learning to Embrace Gerneralization and Personalization on Non-IID Data}

\author{Yushan Zhao, Jinyuan He, Donglai Chen, Weijie Luo, Chong Xie, Ri Zhang, Yonghong Chen and Yan Xu\\
Linklogis Inc.\\
}
\maketitle

\begin{abstract}
   Federated learning (FL) is a decentralized collaborative machine learning (ML) technique. It provides a solution to the issues of isolated data islands and data privacy leakage in industrial ML practices. One major challenge in FL is handling the non-identical and independent distributed (non-IID) data. Current solutions either focus on constructing an all-powerful global model, or customizing personalized local models. Few of them can provide both a well-generalized global model and well-performed local models at the same time. Additionally, many FL solutions to the non-IID problem are benefited from introducing public datasets. However, this will also increase the risk of data leakage. To tackle the problems, we propose a novel data-free distillation framework, Federated Bidirectional Knowledge Distillation (FedBKD). Specifically, we train Generative Adversarial Networks (GAN) for synthetic data. During the GAN training, local models serve as discriminators and their parameters are frozen. The synthetic data is then used for bidirectional distillation between global and local models to achieve knowledge interactions so that performances for both sides are improved. We conduct extensive experiments on 4 benchmarks under different non-IID settings. The results show that FedBKD achieves SOTA performances in every case.\footnote{Code is released at 
\url{https://github.com/FedBKD/FedBKD.git}}
\end{abstract}


\setlength{\intextsep}{0.55cm}
\begin{figure*}[ht]
    \centering
    \includegraphics[scale=0.5]{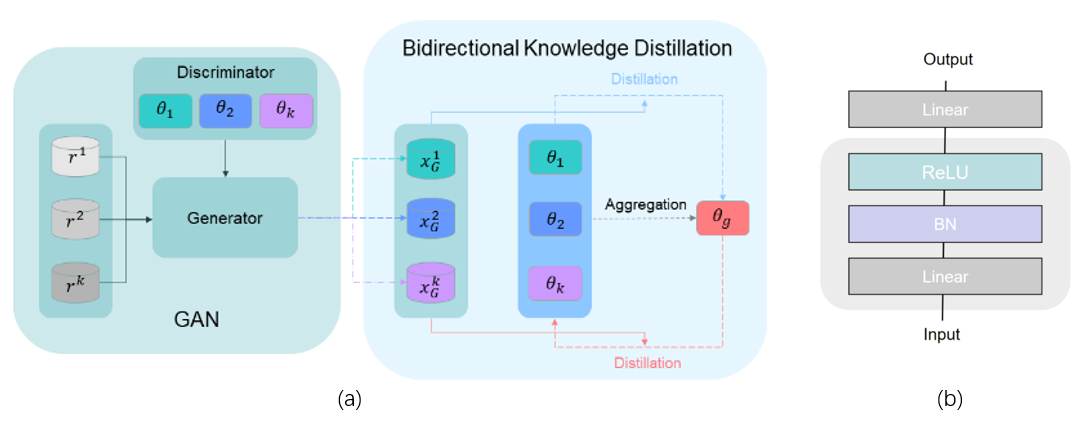}
    
    \caption{Overall framework of FedBKD }
    \label{strcture}
    
\end{figure*}

\section{Introduction}

Federated learning is an emerging technique for decentralized machine learning.
By distributing ML model training across local devices, federated learning increases the flexibility and practical value of machine learning in real world scenarios while minimizing the risk of data privacy leakage \cite{chen2022meta,long2020federated,long2022federated}.

Federated learning (FL) aims to learn a well-generalized global model through multi-round clients’ collaboration.
In each round, the global model is broadcast to every client and fine-tuned with the client's local data. The trained local models are then aggregated to form a new global model. 
Typically, aggregation is achieved by taking the average weights of local models, such as FedAvg \cite{fedavg}. 
This FL paradigm has been proven effective on identical and independently distributed data \cite{chen2021bridging}. 
However, much of the world's data are non-IID \cite{FedAMP,shang2022federated,chen2022personalized,fanprivate}.
The heterogeneity in local data distributions can result in significant weight divergence of the local models even though they have the same initial parameters \cite{hao2021towards}. 
Simple average of the diversified local models will impose a heavy burden on the shared global model to converge at each client and thus worsen the local performances. 
Although some efforts have been made to optimize the global aggregation, such as weighted average \cite{yurochkin2019bayesian,briggs2020federated,wang2020federated}, 
%
model ensemble \cite{guha2019one,shi2021fed,chen2020fedbe}, 
%
and distillation \cite{li2019fedmd,sui2020feded,he2020group,lin2020ensemble},
it is still difficult for a global model to perform well on each client \cite{chen2021bridging}.  

Personalized federated learning (PFL) is a forefront FL paradigm that provides a new idea to cope with the non-IID data \cite{luo2021adapt,tang2022personalized}. 
Instead of learning an all-powerful global model, PFL acknowledges the heterogeneity of clients and customize a particular model for each of them. 
How to selectively inject global information into local models to minimize their empirical risks becomes the major research problem of PFL. 
In literature, regularization and layer substitution are two commonly used strategies for this purpose.
Employment of regularized terms in local training loss will drive updated gradients of local models towards one or more global models. This is helpful to reduce local models' over-fitting on their own data and force them to pay more attention to global information. 
As for layer substitution, networks can be viewed as combinations of representation and classification layers. 
In each local training epoch, representation or classification layers of local models will be substituted by the global ones. 
Layer substitution is a more direct way to pass global information to local models.  

As compared to traditional FL, PFL generally exhibits better local performances \cite{chen2021bridging}. 
However, PFL faces a dilemma: lack of a 'unified' model to provide overview information of all clients. 
Such a model can provide a good initial template for new clients so that their local fine-tuning can be done better and more economically. 
It is especially valuable in many real-world scenarios where client number increases dynamically.
We know that PFL methods also maintain global models, but these global models are `half-done'. 
Most of them are just a simple average of local models' parameters.
They are mainly used to help individual local models to reduce empirical risks. 
They are not specially optimized for direct and fast adaption on new data. 
We will prove this with experiments.

To tackle the above-mentioned problems, we propose a novel data-free distilled federated framework, \textit{Federated Bidirectional Knowledge Distillation (FedBKD)}. 
Specifically, FedBKD is designed for non-IID classification. 
To the best of our knowledge, FedBKD is the first method to use dual distillation to convey knowledge between global and local models. Previous trials are one-way distillation, either from global to local, or vice versa \cite{li2022feddkd,zhu2021data}. 
Under bidirectional distillation, local models have chance to learn from global model to improve their feature extraction ability so that better local performances can be achieved; while global model can integrate knowledge from local models trained on differently distributed data so as to upgrade its generalized ability.
This is how FedBKD embraces generalization and personalization in a single framework. 
Besides, we notice that previous distilled FL methods rely heavily on a public dataset which normally comes from clients' sampled data \cite{li2019fedmd,lin2020ensemble}.
They impose a very strong assumption that clients are willing to contribute to the public dataset. However, this assumption is unrealistic because it leads to a risk of client privacy leakage. 
To guarantee distillation quality while preserving client privacy, FedBKD uses generative networks to synthesize client-alike data. The generative networks are trained against local models (as discriminators), and use maximum response \cite{chen2019data} with diversity regularization \cite{mao2019mode} as training loss to ensure quality of the synthetic data. 
Since the training process does not need any client data, we named the generative network as data-free generator.
We conduct extensive experiments on four benchmarks (CIFAR-10, CIFAR-100, FEMNIST and SENT140) with different non-IID settings. The results show that FedBKD achieves SOTA performance in every experiment.

\section{Related Works}

\subsection{Federated Learning}
Federated learning (FL) aims to learn a well-generalized global model through clients' collaboration. 
FedAvg \cite{fedavg} is the first federated learning method, in which the global model is obtained by simple weighted average of clients' local models. 
However, the weight average method is not suitable for clients with non-IID data, as it will cause serious parameter drifting to the shared global model \cite{li2020federated}. 
Later researches try to improve federated model performances by optimizing the way of clients' knowledge aggregation, includes 
clustering 
\cite{yurochkin2019bayesian,briggs2020federated,wang2020federated}, 
model integration 
\cite{guha2019one,shi2021fed,chen2020fedbe},
and knowledge distillation 
\cite{li2019fedmd,sui2020feded,he2020group,lin2020ensemble}. 
Nevertheless, it is still difficult for a global model to perform well on all clients, especially under the non-IID condition. 
Recently, it is common to synthesize client-alike data to assist direct model training \cite{hao2021towards} or knowledge distillation\cite{li2022feddkd,zhu2021data}. We follow the research trend, but the different thing is that we use synthetic data for dual knowledge distillation, which is proven, in our experiments, to be a more effective way in global-local information transmission.  

\subsection{Personalization Federated Learning}
Personalization federated learning (PFL) \cite{kulkarni2020survey,AvivShamsian2021PersonalizedFL,SaeedVahidian2021PersonalizedFL} is a new federated learning paradigm. It aims to customize a personalized model for each client to cope with the non-IID data.  Unlike the traditional FL paradigm, PFL pay more attention to minimize the empirical risk of client models. Regularization and network layer replacement are two main approaches to realize PFL.  \cite{YutaoHuang2021PersonalizedCF,t2020personalized} use an attention-alike way to achieve clients aggregation, and the aggregated global model is applied as regularization in local model training. 
\cite{LGFEDAVG,LiamCollins2021ExploitingSR,arivazhagan2019federated,chen2021fedmatch} defines a network as representation and classification layers, and replace some of these layers of local models with the global ones. 

\section{Methodology}

FedBKD provides a new paradigm to tackle the non-IID problem. By distillation on high-quality synthetic data, FedBKD enhances local models' feature extraction ability and improves their performances on local data. Meanwhile, it increases global model's generalization performance by allowing it to accumulate experiences on adaption to local data with different distributions.
In this section, we firstly give the task definition of FedBKD. Then we introduce the data-free generator for high-quality synthetic data. The whole algorithm of FedBKD is presented in detailed finally.

\begin{table}[htb]
\begin{tabular}{l}
\hline \textbf{Algorithm 1 FedBKD} \\
\hline 1:  initialize $\theta^{0}$, $G$ at random\\
2:  \textbf{for} $t = 1, 2, \cdots, T$ \textbf{do}\\
3:  \quad $m \leftarrow \max (\gamma \cdot M, 1)$; $S_{t}\leftarrow $random sample $m$ clients,\\ \qquad where $m \in M$\\
4:  \quad \textbf{for} each client $i \in S_{t}$ in parallel \textbf{do}\\
5: \qquad  $\theta_{i}^{t} \leftarrow$ \textbf{ClientUpdate} $\left(\theta_{g}^{t-1}, \hat{\mathcal{X}_{i}}\right)$ // Refer to Sec3.2\\
6: \qquad  Client $i$ sends updated parameter $\theta_{i}^{t}$ to server \\
7:  \quad  \textbf{end for} \\
8:  \quad${x}_{g}^{t} \leftarrow GAN(G, \theta_{s_{t}}^{t})$// Refer to Sec3.1\\
9:  \quad${\theta}_{g}^{t} \leftarrow \frac{1}{m} \cdot \sum_{i \in S_{t}}^{S_{t}} \theta_{i}^{t}$\\
10:  \quad${\theta}_{g}^{'t} \leftarrow global \rightarrow local(\theta_{g}^{t},x_{g}^{st},\theta_{st}^{t})$\\
11:  \quad \textbf{for} each client $i \in S_{t}$ in parallel \textbf{do}\\
12:\quad \quad $\theta_{i}^{'t}$ $\leftarrow local \rightarrow global(\theta_{i}^{t}, x_{G}^{i},\theta_{g}^{t})$ \\
13:\quad \quad $\theta_{i}^{t} \leftarrow \theta_{i}^{'t}$ \\
14:\quad \textbf{return} model parameter $\theta_{i}^{t}$ to client $i$\\
15:\quad \textbf{end for} \\
16: \textbf{end for} \\
17: $\theta_{g}^{t} \leftarrow \theta_{g}^{'t}$ \\
\hline
\end{tabular}
\end{table}

\subsection{Task Definition}
FedBKD is proposed to tackle federated classification. 
The federation scenarios involves one server and $M$ clients, denoted as {$C_1$, $C_2$, ..., $C_M$}. 
The server contains a global model $G$.
Each client has a local model and a local dataset. 
We denote $C_i$'s local model as $\theta_i$, and denote its local dataset as $\mathcal{D}_i=(x_i, y_i)$, where $x_i \in\mathbb{R}^{d}$ represents features, and $y_i \in\mathbb{R}$ represents labels.

We use function $\mathcal{P}(\cdot)$ to denote distribution. The non-IID setting can then be defined as: ${\mathcal{P}(y_i)}\ne\mathcal{P}(y_j)$, where $i, j \in {\{1, \ldots, M\}}$ and $i \neq j$. 

The main objective of FedBKD is to provide customized local models that perform well on their local data, 
while maintaining a well-generalized $G$ that can provide a good initial template to new clients.

\subsection{Data-free Generator}\label{data_gen}
The data-free generator is used to synthesize client-alike data for bidirectional distillation between global and client models.
It distinguishes FedBKD from previous distilled FL methods which rely heavily on clients' data.
With this component, clients' privacy can be preserved better.  
The data-free generator is trained under the GAN framework without  clients' data. 

    

The GAN framework is normally composed of a generator $G$ and a discriminator $D$ \cite{wu2021fedcg}. The discriminator is to distinguish real-world data from data produced by the generator, 
while the generator learns to synthesize more realistic data by considering feedbacks from the discriminator. 
According to a latest research \cite{chen2019data}, a discriminator can be trained in advance. 
That means it is possible to train a generator alone during the adversarial training with the discriminator being frozen.
In FedBKD, we use a client model as the pre-trained discriminator and train a generator to synthesize client-alike data. 
The generator simple and easy to implement, as shown in Fig.\ref{strcture} (b).
The training process of the generator is shown as follow.

Firstly, the generator $G$ is randomly initialized. Let $n$ be the number of synthetic data. The input to $G$ can be denoted as:
\begin{equation}
    R=\left\{r^{1}, r^{2}, \ldots, r^{n}\right\}
\end{equation}
where $r^{n} \in R^{d}$ is a random vector.
The output of $G$ can be noted as:
\begin{equation}
    X_G = \left\{x_{G}^{1}, x_{G}^{2}, \ldots, x_{G}^{n}\right\}, x_{G}^{i}=G\left(r^{i}\right)
\end{equation}
Discriminator $D$ takes $X_{G}$ as input and its output is represented as:
\begin{equation}
    Y_D = \left\{y_{D}^{1}, y_{D}^{2}, \ldots, y_{D}^{n}\right\}, y_{D}^{i}=D\left(x_{G}^{i}\right)
\end{equation}
Since $D$ is a classification model, final class labels of $D$'s input is given by:  
\begin{equation}
t^{i}=\arg \max _{j}\left(y_{D}^{i}\right)_{j}
\end{equation}
The predicted labels of the synthetic data is then given by:
\begin{equation}
    T = \left\{t^{1}, t^{2}, \ldots, t^{n}\right\}
\end{equation}
We consider the synthetic data $x_G^i$ has a good quality if $D(x_G^i)$ is close to one-hot distribution.
The rationale is that such data must exhibit similar patterns with client-data so that the discriminator has great confidence in dealing with it. 
Therefore, we adopt a cross-entropy loss $\mathcal{L}_{o h}$ to optimize the overall quality of synthetic data:

\begin{equation}
\mathcal{L}_{o h}=\frac{1}{n} \sum_{i}^{n} \mathcal{H}_{c r o s s}\left(D(x_G^i), t^{i}\right)
\end{equation}
where $\mathcal{H}_{\text {cross }}$ represents cross-entropy.

Apart from quality control, we also introduce a regularization term \cite{mao2019mode} $\mathcal{L}_{m s}$ to enhance the diversity of the synthetic data:
\begin{equation}
\mathcal{L}_{m s}=-\max _{G}\left(\frac{d\left(G(r^{a}), G(r^{b})\right)}{d\left(r^{a}, r^{b}\right)}\right)
\end{equation}
where $d$ is the distance measurement.

The complete loss function of the data-free generator is given by:
\begin{equation}
\mathcal{L}_{G}=\mathcal{L}_{o h}+\lambda \mathcal{L}_{m s}
\end{equation}
where $\lambda$ is a hyper-parametric weight that used to control the data diversity. Correspondingly, the optimization formula of the data-free generator is represented as:
\begin{equation}
G^{*}=\operatorname{argmin} \frac{1}{n}\left\{\sum_{i=1}^{n} \mathcal{H}_{\text {cross }}\left[\theta_{j}\left(G\left(r_{j}^{i}\right)\right), t_{j}^{i}\right]+\lambda \mathcal{L}_{m s_{j}}\right\}
\end{equation}

\subsection{FedBKD Algorithm}
The federation process of FedBKD is a multi-round learning process.  
Each round consists of a client update stage and a server aggregation stage. 

\textbf{Client Local Update.} 
This stage has two steps: initialization and parameters update. We randomly choose $m$ clients to go through the stage in each communication round. This is a common approach to decrease the communication complexity in federated learning. 

As inspired by FedRep \cite{FedRep}, we define the last network layer of a client model $\theta$ as classification layer $\theta^C$ and all the remaining layers as representation layers $\theta^R$. The classification layer decides the ultimate classification result whereas the representation layers work as feature extractors. 
Let $\theta_{i}(t)$ denotes the parameters of a selected client $i$ in the $t_{th}$ communication round.
$\theta_{i}^R(t)$ is initialized with the representation layers of the global model in the $t-1$ communication round, denoted as $\theta_{g}^R(t-1)$.
Thus, client $i$'s model in the $t$-th communication round is initialized as ${\theta_i}(t)\gets({\theta_g^R}(t-1)\circ{\theta_i^C}(t-1))$. 
Those clients not being selected keep their models remains unchanged.

After initialization, the client model is trained on its own dataset $\mathcal{D}_i$ to update $\theta_i$. We adopt the training technique of FedRep \cite{FedRep} to update each client model by using $\tau + 1$ epochs. Concretely, we freeze the representation layer in the first $\tau$ epochs and only update the classification layer. The process can be represented as:

\begin{equation}
{\theta_i^{C}}(t)=G R D\left(\mathcal{L}_{\mathcal{X}_i}\left({\theta_{i}^{R}}(t-1) \circ {\theta_{i}^{C}}(t-1, \tau)\right), {\theta_{i}^{C}}(t-1, \tau), \alpha\right)
\end{equation}
where $GRD(f,\theta^C,\alpha)$ implicates using function $f$ with stride $\alpha$ to update gradient of classification layer $\theta^C$, and $\mathcal{L}_{\mathcal{X}_i}(\theta)$ denotes the cross-entropy loss of model $\theta$ on $\mathcal{X}_i$. After $\tau$ rounds updates, classification layer of $client_{i}$ can be denoted as ${\theta_{i}^{C}}(t, \tau)$. Meanwhile, fix parameters of classification layers to leverage gradient optimization on representation layers:
\begin{equation}
{\theta_i^{R}}(t)=G R D\left(\mathcal{L}_{\mathcal{X}_i}\left({\theta_{i}^{R}}(t-1) \circ {\theta_{i}^{C}}(t,\tau)\right), {\theta_{i}^{R}}(t-1), \alpha\right)
\end{equation}
The updated local model is finally represented as 
\begin{equation}
{\theta_i}^t=({\theta_{i}^R}(t) \circ {\theta_{i}^C}(t)),
\end{equation}
and $\theta_{i}(t)$ will be transmitted to server for aggregation.

\textbf{Server Aggregation Stage.} Once the server receives the updated local models $\left\{\theta_{1}^{t}, \theta_{2}^{t}, \ldots, \theta_{m}^{t}\right\}$ from the $m$ sampled clients, we update the global model and use the trained generative network to synthesise feature data. The feature data are used for bidirectional knowledge distillation between the global and local models. By feature data, we mean we only synthesize the feature maps that with a similar distribution of the feature maps of real client data. This is to better protect clients' data privacy.

The global model is updated by averaging parameters from the sampled client models:
\begin{equation}
\theta_{g}(t)=\frac{1}{m} \cdot \sum_{i=1}^{m} \theta_{i}(t)
\end{equation}

As for feature data generation, we use the models of sampled clients are used as discriminators in GAN networks to synthesize client-alike feature data $x_{G}^{g}(t)$. More details can be found at the subsection of \textit{Data-free Generator}.  

The knowledge distillation involves two steps.

\textit{1. Global to Local Distillation.} 
Since the global model absorbs rich knowledge from different clients, such distillation helps to improve client models feature extraction ability. 
Specifically, we freeze the classification layer of each local model, and allow each local model using its own synthetic data to successively distill the knowledge of global model's representation layer, the formula of loss function can be represented as:
\begin{equation}
\mathcal{L}_{i \rightarrow g}=K L\left[\theta_{i}^{R}\left(t\right), \theta_{g}^{R}\left(t\right)\right](x_{G}^{i})
\end{equation}
where $KL$ is the abbreviation of Kullback–Leibler divergence, $\mathcal{L}_{i\to{g}}$ represents the loss function of ${\theta_{i}^R}(t)$ distills ${\theta_{g}^R}(t)$. As a result, each local client is updated by:
\begin{equation}
    {\theta_i}(t,n)={\theta_i}(t,n-1)-\alpha \nabla_{{\theta_i}(t,n-1)} \mathcal{L}_{i\to{g}}
\end{equation}
where $n$ represents the distillation is on the $n$-th round.

\textit{2. Local to Global Distillation.} This distillation help to reduce the risk of parameter drift of the global model which is caused by the simple weighted average aggregation method. Besides, it also help to improve the generalization ability of the global model. We will prove it in the \textit{Experiment} section. 
Concretely, we freeze the classification layer of the global model and use the global model to distill the representation layer of corresponding local models on the synthetic data of each client. The loss function is:
\begin{equation}
    \mathcal{L}_{g\to{i}} = kl[{\theta_{g}^R}(t), {\theta_{i}^R}(t)](x_G^i)
\end{equation}
where $\mathcal{L}_{g\to{i}}$ denotes the loss function of ${\theta_{g}^R}^t$ distilling ${\theta_{i}^R}^t$. The update function of global model is:
\begin{equation}
    {\theta_g}(t,n)={\theta_g}(t,n-1)-\alpha \nabla_{{\theta_g}(t,n-1)} \mathcal{L}_{g\to{i}}
\end{equation}
Similarly, we choose to freeze the classification layer ${\theta_{g}^C}(t,n-1)$ when updating the global model to better maintain its generalizability.

\setlength{\intextsep}{0.1cm}
\begin{table*}[ht!]
    \renewcommand\arraystretch{0.9}

    \centering
    \scalebox{1}
    {
    \setlength{\tabcolsep}{10.0 pt}  
    \begin{threeparttable}
    \begin{tabular}{c c c c c c c c}
        \hline 
        \multirow{2}{*}{\textbf{Model}}
        & \multicolumn{3}{c}{\textbf{CIFAR10}} & \multicolumn{2}{c}{\textbf{CIFAR100}} & \multicolumn{1}{c}{\textbf{Sent140}} & \multicolumn{1}{c}{\textbf{FEMNIST}} \\ 
        \cline{2-8}
         & $\mathbf{(100,2)}$ & $\mathbf{(100,5)}$ & 
         $\mathbf{(1000,2)}$ & $\mathbf{(100,5)}$ & $\mathbf{(100,20)}$ & $\mathbf{(183,2)}$ & $\mathbf{(150,3)}$ \\
        \hline
    Local & 89.79 & 70.68 & 78.3 & 75.29 & 41.29 & 69.88 & 60.86       \\
    Fed-MTL & 80.46 & 58.31 & 76.53 & 71.47 & 41.25 & 71.2 & 33.49       \\
    LG-FedAvg & 84.14 & 63.02 & 77.48 & 72.44 & 38.76 & 70.37 & 41.38       \\
    L2GD & 81.04 & 59.98 & 71.96 & 72.13 & 42.84 & 70.67 & 46.56       \\
    APFL & 83.77 & 72.29 & 82.39 & 78.2 & 55.44 & 69.87 & 52.46       \\
    Ditto & 85.39 & 70.34 & 80.36 & 78.91 & 56.34 & 71.04 & 49.63       \\
    FedPer & 87.13 & 73.84 & 81.73 & 76 & 55.68 & 72.12 & 58.53       \\
    FedRep & 87.7 & 75.68 & 83.27 & 79.15 & 56.1 & 72.41 & 59.21       \\
    FedGen & 86.23 & 77.92 & 89.88 & 79.95 & 56.38 & 72.19 & 66.76       \\
    FedRoD & 86.54 & 77.04 & 87.84 & 79.49 & 55.96 & 72.32 & 64.56       \\
    \hline
    \textbf{Ours} & \textbf{88} & \textbf{81.74} & \textbf{90.92} & \textbf{80.56} & \textbf{56.53} & \textbf{74.25} & \textbf{68.55}          \\
    \hline
    FedAvg & 88.68 & 79.31 & 88.94 & 78.46 & 56.75 & 73.83 & 66.73       \\
    FedProx & 88.16 & 80.05 & 89.21 & 77.61 & 55.52 & 74.02 & 66.05       \\
    FedRep & 89.05 & 80.84 & 89.1 & 79.33 & 56.47 & 74.24 & 67.36       \\
    FedRoD & 90.25 & 82.47 & 89.62 & 80.24 & 56.8 & 74.24 & 67.4       \\
    \hline
    \textbf{Ours} & \textbf{91.37} & \textbf{84.36} & \textbf{91.19} & \textbf{81.44} & \textbf{57.02} & \textbf{74.24} & \textbf{69.51}          \\
        \hline 
    \end{tabular}
     
    \end{threeparttable}
    }
    \caption{\centering  Experimental results, where $(x, y)$ denotes  $x$ clients and  $y$ classes per client.}
    \label{mainresults}
\end{table*}

\section{Experiment}

In this section, we conduct extensive experiments to verify the effectiveness of our proposed method FedBKD, and compare it with the state-of-the-art FL algorithms.

\subsection{Benchmark Datasets} 
We evaluate our algorithm on four benchmark datasets: CIFAR-10 \cite{krizhevsky2009learning}, CIFAR-100 \cite{krizhevsky2009learning}, FEMNIST \cite{caldas2018leaf,krizhevsky2009learning} and Sent140 \cite{caldas2018leaf}. Each of CIFAR-10 and FEMNIST contains 10 different categories of images, while CIFAR-100 contains 100 categories of images. Meanwhile, there are 10 different categories of texts in Sent140. Consistent with the settings of LG-FedAvg and FedRep, we manipulate statistical heterogeneity of CIFAR-10 and CIFAR-100 by controlling the number of sample categories $S$ of each client. For FEMNIST, we allocate samples to clients according to the lognormal distribution in [Li et al., 2019], and set the partition of $n=150$ clients, with an average of 148 samples per client. Meanwhile, for Sent140 we consider authors as clients and their tweets as local datasets, thus, employ $n=183$ clients and each client has 72 samples in average.

\subsection{Baselines}
FedAvg\cite{HBrendanMcMahan2017CommunicationEfficientLO} construct a global model through weighted average of local models. 
LG-FedAvg\cite{PaulPuLiang2020ThinkLA} and FedPer\cite{ManojGhuhanArivazhagan2019FederatedLW} are PFL algorithms. The former substitutes classification layers of local models with the global ones, whereas the later replace the representation layers local models with the global ones. 
FedRep\cite{LiamCollins2021ExploitingSR} optimizes the procedures of local training based on FedPer. 
FedProx\cite{li2020federated}, L2GD\cite{FilipHanzely2021FederatedLO} and Ditto\cite{TianLi2021DittoFA} are regularization-based PFL methods, which introduce global regularization in local models training. 
Fed-MTL\cite{CanhTDinh2021FedUAU} regards each client as a unique task to train a single model. 
APFL\cite{YuyangDeng2021AdaptivePF} implements dynamic weighting on local and global models, and analyses convergence in both smooth strongly convex and non-convex settings theoretically. 
FedGen \cite{zhu2021data} synthesises data with global information on server, and add these data to clients' local training to improve personalization ability of local model. 
FedRoD \cite{chen2021bridging} designs two classifiers for client, with one for personalization and the other for generalization purpose, respectively. 

\setlength{\intextsep}{0.1cm}
\begin{table*}[htb!]
    \renewcommand\arraystretch{0.8}
    \centering
    \scalebox{1}
    {
    \setlength{\tabcolsep}{7.8 pt}  
    \begin{threeparttable}
    \begin{tabular}{c c c c c c c c}
        \hline 
        \multirow{2}{*}{\textbf{Model}}
        & \multicolumn{3}{c}{\textbf{CIFAR10}} & \multicolumn{2}{c}{\textbf{CIFAR100}} & \multicolumn{1}{c}{\textbf{Sent140}} & \multicolumn{1}{c}{\textbf{FEMNIST}} \\ 
        \cline{2-8}
         & $\mathbf{(100,2)}$ & $\mathbf{(100,5)}$ & 
         $\mathbf{(1000,2)}$ & $\mathbf{(100,5)}$ & $\mathbf{(100,20)}$ & $\mathbf{(183,2)}$ & $\mathbf{(150,3)}$ \\
        \hline
    FedBKD & 88 & 81.74 & 90.92 & 80.56 & 56.53 & 74.25 & 68.55\\
    with Random Syn Data & 85.58 & 79.25 & 87.64 & 77.54 & 55.1 & 72.19 & 62.73\\
    w/o Distillation & 85.11 & 75.64 & 82.91 & 78.84 & 55.14 & 71.42 & 58.63\\
    w/o Glob to Local DisT & 86.51 & 80.38 & 87.37 & 79.21 & 55.48 & 72.19 & 66.78\\
    w/o Local to Glob DisT & 86.47 & 80.6 & 86.96 & 79.06 & 55.33 & 72.12 & 63.4\\
        \hline 
    \end{tabular}
     
    \end{threeparttable}
    }
    \caption{\centering  Result of ablation experiments}
    \label{Ablation}
\end{table*}

\subsection{Experiment Setup}
The federation process involves 100 communication rounds. For each communication round, we randomly selects 10\% of clients to participate in federated training. In server aggregation stage, we synthesize 5000 feature data for CIFAR-10 $s=20$, 50 feature data for Sent140, 1000 feature data for all other benchmark datasets. In bidirectional distillation, the epochs of \textit{global to local distillation} and \textit{local to global distillation} are set to be 4 and 1, respectively. The distillation learning rate is set to 0.01. In client update stage, client local models are randomly initialized identically, and learning rate is 0.01, following the settings of other methods. We use a batch size of 4 for Sent140, and 10 for all other benchmark datasets. We follow the local training strategy as FedRep \cite{FedRep}, where the representation and classification layers are trained separately. 
The hyper-parameter optimization \cite{bergstra2011algorithms} is conducted to validate the above-mentioned hyper-parameter settings. 

\subsection{Main Result}
In this section, we evaluate both the personalization and the generalization performance of FedBKD on different dataset and under different non-IID settings, with the results being compared to the baselines. 
Specifically, for personalization evaluation, we record the mean prediction accuracy of all client local models on their own test set after each round, and calculate the average of the results of the last 10 rounds. As for generalization evaluation, we retain the global models of the last 5 rounds, and let them to conduct 10 rounds of fine-tuning on a new client that does not participate in the federation. During the process of fine-tuning, we freeze the representation layer of global model and only update the classification layer. The fine-tuned models are then applied on the test set of the new client, and the mean accuracy of the 5 fine-tuned models is used as the generalization metric.

Tab.\ref{mainresults} present the comparative results, with the upper part showing the personalization result and the lower part showing the generalization results. 
It is clear that the proposed framework FedBKD achieves SOTA performances on all conducted experiments.

\subsection{Ablation Study}
In this section, we conduct several experiments to investigate the actual contributions made by the data-free generator and bidirectional distillation to FedBKD.

\textbf{Data-free Generator:} 
We design two comparative experiments for FedBKD. One uses randomly synthetic data for distillation and the other uses the proposed data-free generator. 
Parameters of the two comparative experiments are completely identical. 
The comparative results are illustrated in the first two lines of Tab.\ref{Ablation}. 
It can be observed that, without the proposed data-free generator, FedBKD's performance is visibly worse. The result elucidates the essential of our generative network module. The reason behind will be discussed in the \textit{Discussion} section.

\textbf{Bidirectional Distillation:}  
We design a set of contrast experiments to verify the effectiveness of bidirectional distillation.
\begin{enumerate}
    \item[Exp 1:] Without Distillation. Each sampled client uses its own original local model after each round of federation. The client who does not be sampled uses the averaged representation layer of global model to replace the representation layer of its own local model.
    \item[Exp 2:] Global to Local One-way Distillation. Each sampled client uses the updated local model after each round of federation. The client who does not be sampled is same with Exp 1.
    \item[Exp 3:] Local to global One-way Distillation. Each sampled client is same with Exp 1 while the client who does not be sampled uses the updated representation layer of global model to replace the representation layer of its own local model.
\end{enumerate}

Results are presented in the last three lines of Tab.\ref{Ablation}. The experiment group without distillation has the worst performance. The other two groups with just one-way directional distillation perform better, but the performances are still worse than that of the intact FedBKD. The result indicates the necessity of our designed bidirectional distillation component. The detailed reason is analysed in the \textit{Discussion} section.

\subsection{Discussion}
In this section, we explore how the proposed data-free generator and bidirectional distillation improve FedBKD and the influences of hyper-parameters to the entire framework. 

\begin{figure}[h]
    \centering
    \includegraphics[scale=0.70]{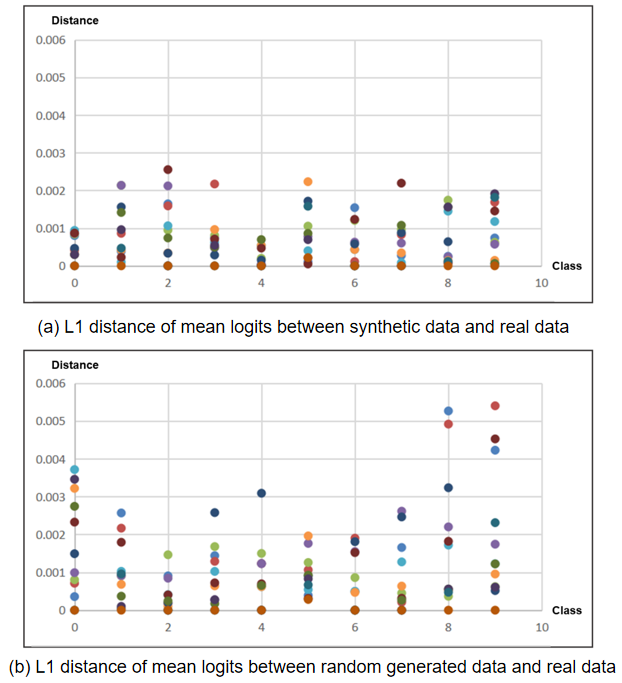}
    \caption{Analytical results for data-free generator}
    \label{GenerativeNetworkAnalytical}
\end{figure}

\textbf{Data-free Generator:} 
The ablation study has shown the proposed data-free generator can improve FedBKD's performance. 
We consider the reason is that the generated feature data is in good quality and has a similar distribution with the feature maps of client's real data. This helps to improve the quality of knowledge distillation and makes global and local models can learn more and useful knowledge from each other. 
To verify this point, we conduct a group of contract experiments: we input the synthesised feature data, randomly generated data, and real client data into the global model, respectively, and record their mean output logits every batch.
For every batch, we then calculate the L1 distance of logits between synthetic data and real data, and between randomly generated data and real data. 
The result is visualized in Fig.\ref{GenerativeNetworkAnalytical}, with each data point representing a L1 distance value. 
The data points in Fig.\ref{GenerativeNetworkAnalytical} (a) are more concentrated and generally lower than those in (b), indicating that the feature maps of the synthetic data can reveal real data patterns. 

\begin{figure}[htb!]
    \centering
    \includegraphics[scale=0.5]{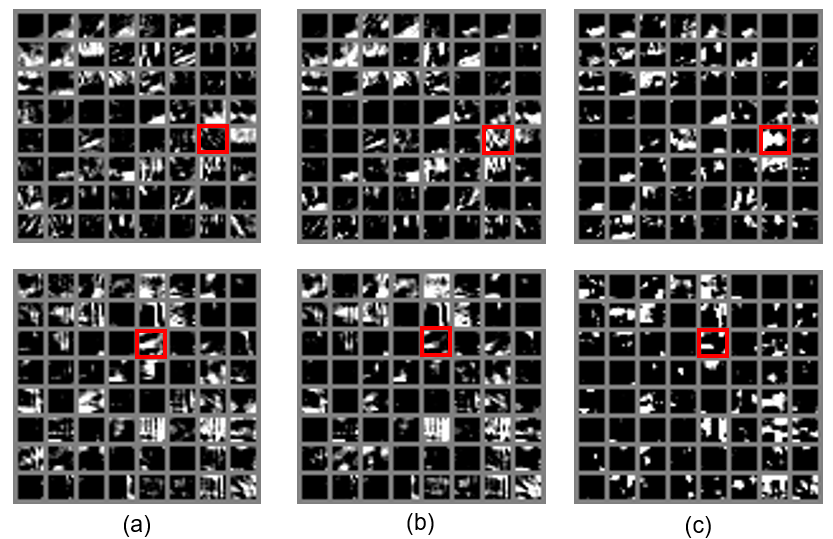}
    \caption{Visual feature maps from three different models. (a) local model before distillation; (b) local model after distillation; (c) upper-bound local model with bi-distillation.}
    \label{synthesisedfeature}
\end{figure}

\textbf{Bidirectional Distillation:} 
To figure out how both global-to-local distillation and local-to-global distillation contribute to the model performance improvement, we conduct an extra experiment. Firstly, during the training of FedBKD, we randomly selected client and save its model parameters before and after global-to-local distillation. We then re-train a standard FedBKD with full data (including training and test data) to work as the upper-bound model. Finally, we give an input the the three models, respectively, and visualize the feature map of a certain convolution layer from the three models, as shown in the upper part of Fig.\ref{synthesisedfeature}.
It can be observed that, compared with Fig.\ref{synthesisedfeature} (a) the visualized feature map shown in Fig.\ref{synthesisedfeature} (b) is closer to Fig.\ref{synthesisedfeature} (c). We highlight three red boxes for reference. This indicates the distilled model accuracy is closer to upper bound of model performance. The situation of global model is similar with that of local model, and the corresponding visualized feature maps are shown in the lower three figures of Fig.\ref{synthesisedfeature}.

\textbf{Effect of Hyper-parameters:}  This section explores the influence of hyper-parameters to FedBKD.

\begin{figure}[h]
    \centering
    \includegraphics[scale=0.3, height=6cm]{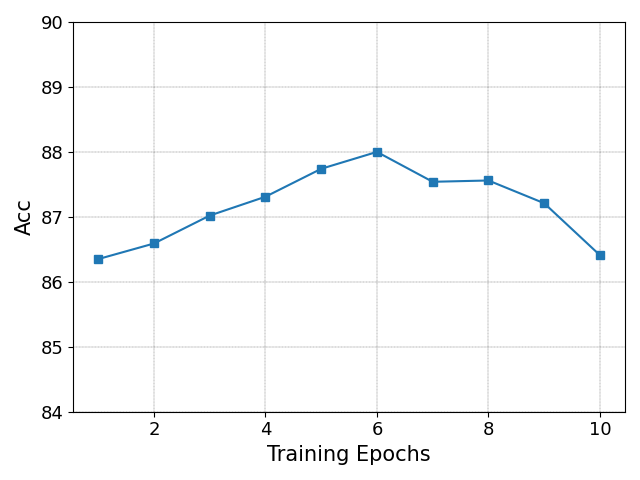}
    \caption{Changing tendency of accuracy according to the training epochs of GAN}
    \label{GAN}
    
\end{figure}

\textit{Training epochs of GAN:} The influence of different training epochs of GAN to experimental results on CIFAR-10 $s=2$ is demonstrated in Fig.\ref{GAN}. 
The ACC is constantly climbing on the early stage, and the model reaches best performance after six epochs then it decreases gradually. 
One possible reason is that, once the number of training epochs beyond certain threshold, it will cause the occurrence of over-fitting in generative network. This can further lead the synthetic data to lose general feature information, thus, decreasing the final effect of distillation. 

\begin{figure}[h]
    \centering
    \includegraphics[scale=0.28, height=6cm]{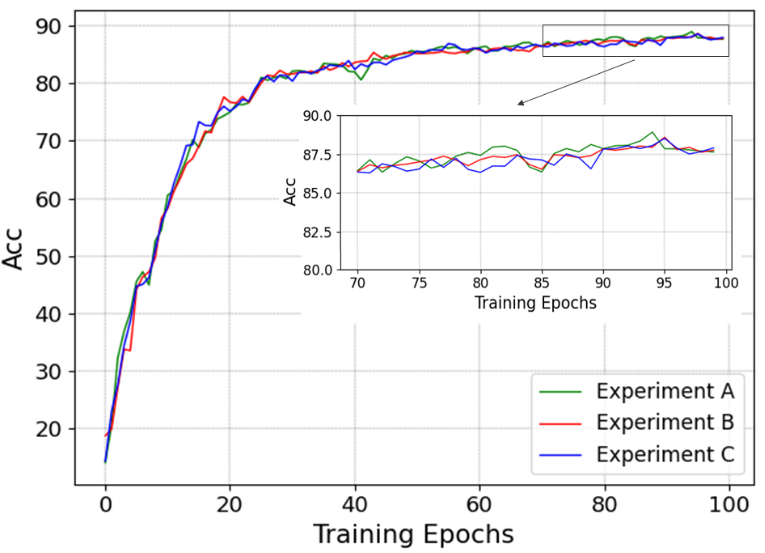}
    \caption{The accuracy of different distillation orders between global and local model}
    \label{global2local}
    
\end{figure}

\textit{Distillation order between global and local models:} In our proposed framework FedBKD, we randomly set the distillation order when global model successively distilling local models and vice versa. 3 different groups of experiment are set to discuss the effect of distillation order. Specifically, apart from the order of distillation, the remaining experimental variables are identical in the 3 groups. From Fig.\ref{global2local}, we note that the influence of distillation order to the final experimental result is negligible low. This is because, during local-to-global distillation, each local model can be regarded as a batch to global model. The situation is similar in global-to-local distillation. As for model training in deep learning, adjusting the input batch order in each training round has negligible effect to final model performance, which is similar to the distillation order between global and local models.

\section{Conclusion}
In this work, we propose a novel distilled federated learning framework, FedBKD, for non-IID classification. 
To the best of our knowledge, FedBKD is the first attempt to use bidirectional-distillation to improve knowledge transmission between global and local models. 
With bidirectional-distillation, local models enhance their feature extraction ability and achieve better performances on their local data, while global model accumulates more experiences on adaption to data with different distributions and increases its generalization ability. 
Besides, to avoid clients' data privacy leakage and meanwhile ensure the distillation quality, FedBKD customizes a data-free generator to synthesize high-quality client-alike data. 
Extensive experiments have been done on four benchmark datasets (CIFAR-10, CIFAR-100, FEMNIST and SENT140) with different non-IID settings. The results have demonstrated the effectiveness of FedBKD as it achieves SOTA performance on every experiment. 
The significance of bidirectional-distillation and data-free generator has also been proven in the experiments. 
Currently, FedBKD can only work for non-IID classification. 
In the future, we will extend FedBKD's capability to cover more complicated machine learning tasks. 

\appendix

{\small
\bibliographystyle{ieee_fullname}
\bibliography{egbib}
}

\end{document}